\documentclass[11pt,a4paper]{article}
\pdfoutput=1
\PassOptionsToPackage{hyphens}{url}\usepackage[hyperref]{acl2021}
\usepackage{times}
\usepackage{latexsym}

\usepackage{microtype}

\usepackage{xurl}
\usepackage{booktabs}
\usepackage{amssymb}
\usepackage{cleveref}
\usepackage{xspace}
\usepackage{xcolor,soul}

\usepackage{tikz}
\usepackage[german,brazilian]{babel}

\newcommand{\datasetName}{{\sc X-Fact}\xspace}

\crefformat{section}{\S#2#1#3}  
\definecolor{orange2}{rgb}{0.95,0.35,0}

\def\cmark{\tikz\fill[scale=0.4](0,.35) -- (.25,0) -- (1,.7) -- (.25,.15) -- cycle;} 

\newcommand{\alphaOne}{$\alpha_1$\xspace}
\newcommand{\alphaTwo}{$\alpha_2$\xspace}
\newcommand{\alphaThree}{$\alpha_3$\xspace}

\aclfinalcopy % Uncomment this line for the final submission
\def\aclpaperid{4130} %  Enter the acl Paper ID here

\title{\datasetName: A New Benchmark Dataset for Multilingual Fact Checking}

\author{Ashim Gupta, Vivek Srikumar \\
  School of Computing, \\ University of Utah \\
  \texttt{\{ashim,svivek\}@cs.utah.edu}\\}

\date{}

\begin{document}
\maketitle

\begin{abstract}
In this work, we introduce \datasetName: the largest publicly available multilingual dataset for factual verification of naturally existing real-world claims. The dataset contains short statements in 25 languages and is labeled for veracity by expert fact-checkers. The dataset includes a multilingual evaluation benchmark that measures both out-of-domain generalization, and zero-shot capabilities of the multilingual models. Using state-of-the-art multilingual transformer-based models, we develop several automated fact-checking models that, along with textual claims, make use of additional metadata and evidence from news stories retrieved using a search engine.
Empirically, our best model attains an F-score of around 40\%, suggesting that our dataset is a challenging benchmark for evaluation of multilingual fact-checking models.
% Our experiments and analysis suggest that although more sophisticated evidence-based models marginally outperform claim-only models, their applicability to automated fact checking is severely handicapped by their inability to process large documents.
\end{abstract}
\section{Introduction}
\label{sec:intro}

% \input{tables/examples}
% The spread of fake news and misinformation on the web has increased demand for manual verification 
Curbing the spread of fake news and misinformation on the web has become an important societal challenge.  Several fact-checking initiatives, such as  PolitiFact,\footnote{\url{https://www.politifact.com/}} expend a significant amount of manual labor to investigate and determine the truthfulness of 
% this information. 
viral statements made by public figures, organizations, and social media users.
% or viral claims made by users on social media portals.
Of course, since this process is time-consuming, often, a large number of falsified statements go unchecked. 

% Mention somewhere that other languages should at least get equal attention, since large number of social media users converse in non-english.

% Concurrently, 
With the aim of assisting fact-checkers, researchers in NLP have sought to develop computational approaches to fact-checking~\citep{vlachos-riedel-2014-fact,wang-2017-liar,perez-rosas-etal-2018-automatic}. Many such works use the FEVER dataset, which contains claims extracted from Wikipedia documents~\citep{thorne-etal-2018-fever}. Using real-world claims, 
~\citet{wang-2017-liar} introduced LIAR, a dataset with 12,836 claims from PolitiFact. Recently,~\citet{augenstein-etal-2019-multifc} introduced MultiFC, an even larger corpus of 34,918 claims collected from 26 fact-checking websites.
% , again in English. 

Although misinformation transcends countries and languages~\citep{bradshaw2019global,islam2020covid}, much of the recent work focuses on claims and statements made in English.
% Notably, much of the recent work focuses on claims and statements made in English, although the issue of misinformation transcends countries and languages~\citep{bradshaw2019global,islam2020covid}.
Developing Automated Fact Checking (AFC) systems in other languages is much more challenging, the primary reason being the absence of a manually annotated benchmark dataset for those languages. 
Moreover, there are fewer fact-checkers in these languages, and as a result, a non-English monolingual dataset  will inevitably be small and less effective in developing fact-checking systems. 
As recent research points out, a possible solution in dealing with data scarcity is to train multilingual models~\citep{aharoni-etal-2019-massively,wu-dredze-2019-beto,hu2020xtreme}.
% Moreover, recent research shows that multilingual models are more effective than monolingual models in resource scarce scenarios~\citep{aharoni-etal-2019-massively,wu-dredze-2019-beto,hu2020xtreme}.  
% Indeed, this observation motivates the multilingual nature of our work. 
Indeed, this finding motivates us to construct a large multilingual resource that the research community can use to further the development of fact-checking systems in languages other than English.
% Moreover, collecting separate dataset for each language is ineffective, as fact checks done in other languages 
% Collecting such a dataset also presents a formidable challenge:
% A major bottleneck for development of automated fact checking systems for other languages is the lack of availability of a manually labeled benchmark dataset. 

% We aim to fill this gap by contributing a large multilingual dataset of short statements manually labeled for their veracity by expert journalists and fact checkers. 

% Several fact checking initiatives such as politifact.com 
% This has led to the establishment of several fact checking websites and portals such as politifact.com. Typically, these organizations employ 
% Several fact checking  

Recent efforts in the construction of a multilingual dataset are limited, both in scope and in size~\cite{shahifakecovid,patwa2020fighting}. For instance, FakeCovid, a dataset introduced by~\citet{shahifakecovid} contains 3066 non-English claims about COVID-19. 
In comparison, \datasetName\,contains 31,189 general domain non-English claims from 25 languages.
Moreover, FakeCovid contains only two labels, namely, \textit{False}, and \textit{
Others}. We argue that this is undesirable, as fact checking is a fine-grained classification task.
% with a graded notion of truthfulness. Most statements are neither entirely true nor entirely false~\cite{rashkin2017truth}.  
Due to subtle differences in language, most claims are neither entirely true nor entirely false~\cite{rashkin-etal-2017-truth}.
In contrast, our dataset contains seven labels---we make distinctions between \textit{true}, \textit{mostly true}, \textit{half-true} etc.~\Cref{tab:examp_claim} shows two such examples from German and Brazilian Portuguese.

In summary, our contributions are:
\begin{enumerate}
\item We release a multilingual fact-checking benchmark \datasetName 
, which includes 31,189 
% \ag{check number?} 
short statements labeled for factual correctness and covers 25 typologically diverse languages across 11 language families.
% , annotated for their factual correctness by expert journalists. 
\datasetName\, is an order of magnitude larger than any other multilingual dataset available for fact checking. 

\item Apart from the standard test set, we create two additional challenge sets to evaluate fact checking systems' generalization abilities across different domains and languages.

\item We report results for several modeling approaches and find that these models underperform on all three test sets in our benchmark, suggesting the need for more sophisticated and robust modeling methods.
\end{enumerate}

The \datasetName dataset, and the code for our experiments, can be obtained at \url{https://github.com/utahnlp/x-fact}.

\section{The \datasetName\,Dataset}
\label{sec:dataset}
\begin{table}[t]
\fontsize{10}{10}\selectfont
\begin{tabular}{p{1.8cm}p{5.0cm}}   %{@{}ll@{}} {\textwidth}
\toprule
% Feature & Value \\ 
\midrule
\textbf{Claim} & \textit{Muslimische Gebete sind Pflichtprogramm an katholischer Schule.} \\ 
 & Muslim prayers are compulsory in Catholic schools. \\
Label & Mostly-False (\textit{Gr\"osstenteils Falsch})\\
Claimant & Freie Welt \\
Language & German\\
Source &  \scriptsize{\url{de.correctiv.org}}\\
Claim Date & March 16, 2018 \\
Review Date &  March 23, 2018\\
\midrule
\textbf{Claim} & \textit{Temos, hoje, a despesa de Previdência Social representando 57\% do orçamento.} \\ 
 & Today, we have Social Security expenses representing 57\% of the budget. \\
Label & Partly-True (\textit{Exagerado})\\
Claimant & Henrique Meirelles \\
Language & Portuguese (Brazilian)\\
Source &  \scriptsize{\url{pt.piaui.folha.uol.com.br}}\\
Claim Date & None \\
Review Date &  May 2, 2018\\
\midrule
\bottomrule
\end{tabular}
\caption{\label{tab:examp_claim} Examples from \datasetName. Original labels are shown in parenthesis along with the manually mapped labels. For reference, translations are also shown.}
\end{table}
% Collecting a large multilingual dataset of claims presents its own challenges. 
% This section describes data collection in terms of selecting sources of fact checks, preprocessing, and filtering of the dataset.

\datasetName\,is constructed from several fact-checking sources. We briefly outline this process here.

% \subsection{Dataset Construction}

\paragraph{Sources of Claims. } We relied on a list of non-partisan fact-checkers compiled by International Fact-Checking Network (IFCN)\footnote{\url{https://www.poynter.org/ifcn/}}, and Duke Reporter's Lab\footnote{\url{https://reporterslab.org/fact-checking/}}. We removed all the websites that conduct fact-checks in English and are covered by previous work\cite{wang-2017-liar,augenstein-etal-2019-multifc}. As a starting point, we first queried Google's Fact Check Explorer (GFCE)\footnote{\url{https://toolbox.google.com/factcheck/explorer}}
for all the fact-checks done by a particular website. Then we crawled the linked article on the website and additional metadata such as claimant, URL, date of the claim. For websites not linked through GFCE, we directly crawled all the available fact-checking articles from the fact-checker's website. We left out some fact-checkers because either the claims on their websites were not well specified or the fact-checker did not use any rating scale. We performed semi-automated text processing to remove duplicate claims and examples where the label appeared in the claim itself. 
% Finally, we obtained data from a total of 85 fact checkers for further processing.
This resulted in data from a total of 85 fact checkers for further processing.
Refer to the appendix for more details on the this process.

\paragraph{Filtering the Dataset. } There are two major challenges in using the crawled data directly: a) the labels are in different languages, and b) each fact checker uses a different rating scale for categorization. To deal with these issues, first, we manually translated all ratings to English, followed by semi-automatic merging of labels if they were found to be synonyms. Second, in consultation with Factly,\footnote{\url{https://factly.in}} an IFCN signatory, we created a rating scale compatible with most fact-checkers. Our label set contains five labels with a decreasing level of truthfulness: \textit{True}, \textit{Mostly-True}, \textit{Partly-True}, \textit{Mostly-False}, and \textit{False}. To encompass several other cases where assigning a label is difficult due to lack of evidence or subjective interpretations, we introduced \textit{Unverifiable} as another label. A final label \textit{Other} was used to denote cases that do not fall under the above-specified categories. Following the process described, we reviewed each fact-checker's rating system along with some examples and manually mapped these labels to our newly designed label scheme. See~\cref{tab:examp_claim} for examples. In our subsequent discussions, we refer to each fact-checking website as a \textit{source}. 
% Supplemental provides additional discussion on this procedure~\ag{appendix cited}.

We found that the data from several sources was dominated by a single label ($> 80\%$). Since it is difficult to train machine learning models on highly imbalanced datasets, we removed 54 such websites.
% from our consideration. 
% In total, this resulted in 
% % 46168 - 12310
% 33,858 claims. 
We additionally removed fact-checking websites that contained fewer than 60 examples. 
% Finally, our dataset contains a total of 31,723 fact checks. 
In total, our dataset contains 31,189 fact-checks. 
% In our subsequent discussions, we refer to each fact checking website as a \textit{source}.

% \begin{table}
%   \centering
%   \begin{tabular}{lcc}
%     \toprule
%     Data split       & \# claims & \# languages \\
%     \midrule
%     Train            & 18030        & 12       \\
%     Dev              & 2396        & 12       \\
%     In-Domain Test   & 3616        & 12       \\
%     Out-of-Domain Test   &         & 12       \\
%     Zero-Shot Test & 3381        & 13       \\  
%     \bottomrule
%   \end{tabular}
%   \caption{Number of claims and languages for each data split}
%   \label{tab:data-splits-stats}
% \end{table}
\begin{table}
  \centering
  \begin{tabular}{lrr}
    \toprule
    Data split       & \# claims & \# languages \\
    \midrule
    Train            & 19079        & 13       \\
    Development              & 2535        & 12       \\
    In-domain (\alphaOne)   & 3826        & 12       \\
    Out-of-domain (\alphaTwo)   &     2368    & 4       \\
    Zero-Shot (\alphaThree) & 3381        & 12       \\  
    \bottomrule
  \end{tabular}
  \caption{Dataset details. \datasetName contains three challenge sets, namely, In-domain Test (\alphaOne), Out-of-domain Test (\alphaTwo), Zero-Shot Test (\alphaThree). }
  \label{tab:data-splits-stats}
\end{table}

% \begin{table}
%   \centering
%   \begin{tabular}{lcc}
%     \toprule
%     Data split       & \# claims &  languages \\
%     \midrule
%     Train            & 18030        & Arabic, Hindi, Tamil, Spanish,       \\
%     Dev              & 2396        & 12       \\
%     \alphaOne   & 3616        & 12       \\
%     \alphaTwo   &         & 12       \\
%     \alphaThree & 3381        & 13       \\  
%     \bottomrule
%   \end{tabular}
%   \caption{Number of claims and languages for each data split}
%   \label{tab:data-splits-stats}
% \end{table}
% \subsection{The Need for Multi-Faceted Evaluation}
% \paragraph{The Need for Multi-Faceted Evaluation.}
\paragraph{A Single Test Set is Not Sufficient.}
Recent advances in NLP have shown that multilingual models are effective for cross-lingual transfer~\citep{kondratyuk-straka-2019-75,wu-dredze-2019-beto,hu2020xtreme}. A multilingual fact-checking system of similar transfer capabilities will certainly be an asset, especially in languages with no or few fact-checkers. From this perspective, we seek to provide a robust evaluation benchmark that can help us understand the generalization abilities of our fact-checking systems.

With this objective, we construct three test sets, namely \alphaOne, \alphaTwo, and \alphaThree.\footnote{The names for our test sets, and the idea of having multiple test sets without corresponding training sets, is inspired by~\citet{gupta-etal-2020-infotabs}.} The first test set (\alphaOne) is distributionally similar to the training set. The \alphaOne\,set contains fact-checks from the same languages and sources as the training set.

Second, the \textit{out-of-domain} test set (\alphaTwo), contains claims from the same languages as the training set but are from a different source. A model that performs well on both \alphaOne\,and \alphaTwo\, can be presumed to generalize across different source distributions. 
% in that language.

Third test set is the \textit{zero-shot} set  (\alphaThree), which seeks to measure the cross-lingual transfer abilities of fact-checking systems. The \alphaThree\,set contains claims from languages not contained in the training set. Models that overfit language-specific artifacts will underperform on \alphaThree.

% \paragraph{Languages } The twelve languages with most labeled examples are used to construct training, development, and \alphaOne\, test set. 
% % On average, the training set contains 1338 examples from each language - the smallest being Serbian with . 
% Among these languages, average number of examples per language is 1784, with smallest being Serbian with 835 examples.
% % To construct the training set, we select websites with at least 900 fact checks. 
% For each of these languages, we split the data into training set (75\%), development set (10\%), and \alphaOne\,test set (15 \%). From the remaining set of websites, we separate out those in languages that are not present in the training set and construct \alphaThree. The remaining websites are used to construct out-of-domain evaluation set - \alphaTwo. See~\cref{tab:data-splits-stats} for dataset details. 

\paragraph{Languages.}
% The twelve languages with most labeled examples are used to construct training, development, and \alphaOne\, test set. 
For training and development, we choose the top twelve languages based on the number of labeled examples.
The average number of examples per language is 1784, with Serbian being the smallest (835). We split the data into training (75\%), development (10\%), and \alphaOne\,test set (15\%). This leaves us with 13 languages for our zero-shot test set (\alphaThree). The remaining set of sources form our out-of-domain test set (\alphaTwo). See~\cref{tab:data-splits-stats} for the number of claims and langauges in each of these splits.

In total, \datasetName\, covers the following 25 languages (shown with their ISO 639-1 code for brevity): ar, az, bn,  de, es, fa, fr, gu, hi, id, it, ka, mr, no, nl, pa, pl, pt, ro, ru, si, sr, sq, ta, tr. Please refer to the appendix for more details.

\section{Experiments and Results}
\label{sec:results}
\subsection{Experimental Setting}
\label{sec:models}
The goal of our experiments is to study how different modeling choices address the task of multilingual fact-checking. All our experiments use mBERT, the multilingual variant of BERT~\cite{devlin-etal-2019-bert} and use macro F1 score as the evaluation metric.\footnote{Although it is possible to develop partial scoring metrics, which we leave for future work to explore.} We report average F1 scores and standard deviations on four runs with different random seeds. 

We implement the following multilingual models as baselines for future work:

\begin{enumerate}
    \item \textbf{Claim Only Model (Claim-Only): } We provide textual claim as the only input to the model, in effect treating the problem as a simple sentence classification problem. 
    % Notice that this setting is similar to that of hypothesis only models for the task of NLI~\cite{poliak2018hypothesis}. 
    
    \item \textbf{Attention-based Evidence Aggregator (Attn-EA): } Typically, to determine the veracity of a claim, fact-checkers first gather relevant evidence by performing a web search and then aggregate this evidence to reach their final decision. We emulate this procedure by developing an attention-based evidence aggregation model that operates on evidence documents retrieved after performing web search with the claim using Google. For each claim, we obtain the top five results and use them as evidence. Using full text from web pages is not feasible, as the mBERT model has a restricted input sequence length of 512. Following previous work~\cite{augenstein-etal-2019-multifc}, we use snippets from search results as our  evidence.
    
    For a given claim and a collection of $n$ evidence documents, we first encode the claim and evidences separately using mBERT by extracting the output of the \texttt{CLS} token, denoted as: $\mathbf{c}$, $[\mathbf{e_1}, \mathbf{e_2}, ..., \mathbf{e_n}]$. We first apply dot-product attention~\cite{luong-etal-2015-effective} to obtain the attention weights $[\alpha_1, \alpha_2, ..., \alpha_n]$, and then compute a linear combination using these attention coefficients:
    $\mathbf{e} = \sum_{i}\alpha_i \mathbf{e_i}$. This representation is then concatenated with $\mathbf{c}$ and fed to the classification layer. In all our experiments, we fix the number of evidence documents to five.
    
    \item \textbf{Augmenting metadata (+Meta): } We concatenate additional key-value metadata with the claim text by representing it as a sequence of the form:  \texttt{Key} \texttt{:} \texttt{Value}~\cite{chen2019tabfact}. This metadata includes the \texttt{language}, \texttt{website-name}, \texttt{claimant}, \texttt{claim-date}, and \texttt{review-date}. If a certain field is not available for a claim, we represent the value by \texttt{none}.
\end{enumerate}

All the models are trained in a multilingual setting, i.e., a single model is trained for all languages. We could not use monolingual models as the trained monolingual models were unstable due to the small size of data for each language.
% We could not train stable monolingual models for every language due to the small size of data for each language.

% Only other multilingual pre-trained model is XLM-RoBERTa~\cite{conneau2020unsupervised} -  
% \paragraph{Evaluation Metrics. }
% We use the Macro F1 score and a new metric that acknowledges the graded notion of truthfulness and assigns partial scores to classification decisions. The intuition here is that a model that labels a 'true' example as 'mostly true' should be penalized less than the one that predicts 'false'. The corresponding partial score matrix is presented in the supplemental. In all subsequent results, we report an average of F1 and Partial Score. Detailed results are present in the supplemental.

\begin{table}[h]
\centering
\resizebox{0.99\columnwidth}{!}{
\begin{tabular}{lrrrr}
  \toprule
  Model &  \alphaOne & \alphaTwo & \alphaThree \\
  \midrule
  Majority & 6.9\scriptsize\textcolor{gray}{(--)}  & 10.6\scriptsize\textcolor{gray}{(--)} & 7.6\scriptsize\textcolor{gray}{(--)}\\
  Claim-Only        & 38.2\scriptsize\textcolor{gray}{(0.9)}   & \textbf{16.2}\scriptsize\textcolor{gray}{(0.9)}   & 14.7\scriptsize\textcolor{gray}{(0.6)}   \\
  Claim-Only + Meta     & 39.4\scriptsize\textcolor{gray}{(0.9)}  &  15.4\scriptsize\textcolor{gray}{(0.8)} &  \textbf{16.7}\scriptsize\textcolor{gray}{(1.1)} \\ 
    \midrule
  Attn-EA (Random)        &  37.5\scriptsize\textcolor{gray}{(0.8)} & 16.3\scriptsize\textcolor{gray}{(0.5)}  & 14.9\scriptsize\textcolor{gray}{(1.2)}  \\ 
  Attn-EA                 & 38.9\scriptsize\textcolor{gray}{(0.2)}  & 15.7\scriptsize\textcolor{gray}{(0.1)}  & 16.5\scriptsize\textcolor{gray}{(0.7)}  \\ 
  Attn-EA + Meta          &  \textbf{41.9}\scriptsize\textcolor{gray}{(1.2)}  & 15.4\scriptsize\textcolor{gray}{(1.5)}  &  16.0\scriptsize\textcolor{gray}{(0.3)} \\
  \bottomrule
\end{tabular}
}
\caption{Average F1 scores (and standard deviations) of the models studied in this work. Models in top rows are claim-only models while those in bottom are evidence-based. Attn-EA (Random) denotes the results of the evidence-based model when it is trained with random search snippets. (+ Meta) models denote those augmented with additional metadata.}
\label{tab:main_results}
\end{table}

% \begin{table}[h]
% \centering
% \begin{tabular}{ccccc}
%   \toprule
%   Model &  \alphaOne & \alphaTwo & \alphaThree \\
%   \midrule
%   Claim Only & 60.6  & 45.8  & 45.8  \\
%   Claim + Meta     & 63.4  & 49.6  &  50.4 \\ 
%     \midrule
%   Attn-Doc (Random)        & 62.7  & 50.5  & 50.8  \\ 
%   Attn-Doc                 & 62.7  & 50.5  & 50.8  \\ 
%   Attn-Doc + Meta          &  60.4  & 48.6  &  48.8 \\
%   \bottomrule
% \end{tabular}
% \caption{Accuracy of hypothesis-only baselines on the \datasetName\, Dev and test sets}
% \label{tab:main_results}
% \end{table}

\subsection{Results}
The results are shown in~\cref{tab:main_results}. We will discuss results by answering a series of research questions. As an indicator of label distribution, we include a majority baseline with the most frequent label of the distribution (i.e. \texttt{false}). 

\paragraph{Does the dataset exhibit claim-only bias? } 
Before moving to more sophisticated systems, let us first examine if the model can predict a statement's veracity by only using the textual claim. Note that this setting is similar to that of hypothesis only models for the task of Natural Language Inference (NLI)~\cite{poliak-etal-2018-hypothesis}. From~\cref{tab:main_results}, we see that a claim-only model outperforms a majority baseline by a large margin. We can draw two inferences: a) A significant number of examples in \alphaOne\,can be labeled by just relying on the textual claim, and b) the claim-only model has learned spurious correlations from the dataset.

\paragraph{Do search snippets improve fact-checking?}
First, results from~\cref{tab:main_results} show that augmenting models with metadata is helpful. Second, using search snippets as evidence with an attention-based model along with metadata improves performance by 2.5 percentage points on the in-domain test set (\alphaOne). 
% Another approach to determine this is to pair a claim with a random snippet of same language. 
To further validate that snippets indeed help the evidence-based model, we perform another experiment in which we pair each claim with random search snippets of the same language. Since there is no relevant evidence, the performance is indeed similar to the claim-only model. This again confirms our finding that the dataset exhibits some claim-only bias. 
% We see that the performance drops significantly 

While the Attn-EA model provides some performance improvement on the in-domain test set, surprisingly, the claim-only model outperforms the evidence-based model by a small margin on \alphaThree. 
This might be due to the evidence-based over-fitting the in-domain data. 

\paragraph{How informative are the search snippets?}
Note that we used snippets to summarize the retrieved search results.
To gauge the relevance of these snippets, we manually examine 100 examples from \alphaOne\,test set for Hindi. 
Our preliminary analysis reveals that only 45\% of snippets provide sufficient information to classify the claim, indicating why the performance increase with the evidence-based model is small. 
Our same analysis suggests that for 83\% of the examples, using full text of the web pages provides sufficient evidence to determine veracity of the claim. Hypothetically, this means, were the models able to ingest large documents (web pages), their performance increase could have been much more significant. 

\paragraph{Do the models generalize across sources and languages?} 
We observe that performance on \alphaTwo\,and \alphaThree\,is worse than on \alphaOne, not only highlighting the difficulty of these challenge sets, but also showing that models overfit both source-specific patterns (\alphaTwo) and language-specific patterns (\alphaThree).
% A significant drop in performance on \alphaTwo, and \alphaThree, for both Claim-only model and the Attn-EA model highlights 
% the poor generalizability of these multilingual models. We can conclude that all of the models learn language-specific as well as source-specific characteristics. 

Importantly, these results underscore the utility of our challenge sets in assessing model generalizability as well as diagnosing overfitting.

\begin{table}[h]
\centering
\resizebox{0.99\columnwidth}{!}{
\begin{tabular}{lrrr}
  \toprule
  Model &  \alphaOne & \alphaTwo & \alphaThree \\
  \midrule
  \multicolumn{4}{c}{\datasetName}        \\ 
  Claim-Only + Meta     & 39.4\scriptsize\textcolor{gray}{(0.9)}  &  \textbf{15.4}\scriptsize\textcolor{gray}{(0.8)} &  \textbf{16.7}\scriptsize\textcolor{gray}{(1.1)} \\ 
  Attn-EA + Meta          & \textbf{41.9}\scriptsize\textcolor{gray}{(1.2)}  & \textbf{15.4}\scriptsize\textcolor{gray}{(1.5)}  &  16.0\scriptsize\textcolor{gray}{(0.3)} \\
    \midrule
    \multicolumn{4}{c}{\datasetName\,+ English}        \\ 
  Claim-Only + Meta     & 37.1\scriptsize\textcolor{gray}{(2.7)}  &  14.5\scriptsize\textcolor{gray}{(0.5)} &  14.4\scriptsize\textcolor{gray}{(0.3)} \\ 
  Attn-EA + Meta          &  38.0\scriptsize\textcolor{gray}{(4.5)}  & 14.7\scriptsize\textcolor{gray}{(2.6)}  &  14.3\scriptsize\textcolor{gray}{(1.9)} \\
  \bottomrule
\end{tabular}
}
\caption{Performance comparison when augmenting the dataset with 12,311 English claims from PolitiFact. Average F1 scores (and standard deviations) of the models are reported over four random runs.}
\label{tab:politifact_results}
\end{table}
\paragraph{Can we improve performance by augmenting training data with English claims?}
Since \datasetName\, does not contain any examples from English, 
% Could data augmentation from English help? 
we answer this question by augmenting the training set with 12,311 claims from the PolitiFact subset of the MultiFC~\cite{augenstein-etal-2019-multifc}. 
Results are shown in~\cref{tab:politifact_results}. Interestingly, we see that augmenting the models with English data hurts model performance. 
% While this finding might seem counter intuitive, a closer look at the augmented
A possible cause is that the augmented data mostly contains political claims, while our dataset contains general claims.

% \paragraph{Metadata is important}

% \paragraph{}
% \paragraph{Discussion}
% Broadly, our results show that: 1) \datasetName\, contains some amount of claim-only bias which all of the multilingual models learn; and 2) model performances drop on \alphaTwo\,and \alphaThree, suggesting that state-of-the-art models fail to generalize across sources and languages, indicating the need for new modeling strategies.

% \paragraph{How do the generalization?}
% Importantly, results on \alphaTwo, and \alphaThree\, indicate that model shows poor generalization abilities for both out-of-domain setting, zero-shot cross lingual transfer.
% \paragraph{Does using additional data from English help?}
% Mention why not to use crowd sourcing for fact checking
% \paragraph{Discussion. }

\section{Conclusion}
We presented \datasetName, the currently
largest multilingual dataset for fact-checking. Compared to the prior work, \datasetName\,is an order of magnitude larger, enabling the exploration of large transformer-based multilingual approaches to fact-checking. We presented results for several multilingual modeling methods and showed that the models find this new dataset challenging.
We envision our dataset as an important benchmark in development and evaluation of multilingual approaches to fact-checking.

\section*{Acknowledgments}
We would like to thank members of the Utah NLP group for their valuable insights, reviewers for their helpful feedback, and the team of Factly,\footnote{\url{https://factly.in/}} especially Mr. Shashi Kiran Deshetti, for discussions in developing a rating scale compatible with most fact-checkers.
The authors acknowledge the support of NSF grants \#1801446 (SATC) and \#1822877 (Cyberlearning) and an award from Verisk Inc.

\bibliographystyle{acl_natbib}
\bibliography{anthology,acl2021}

\clearpage
\appendix
\section{Details on Dataset Construction}
\begin{enumerate}

\item As mentioned in the paper, we omit several fact-checking websites from our data. A large number of these websites are not amenable to crawling and scraping the data. For instance, AFP\footnote{\url{https://factuel.afp.com/}} is a prominent fact-checker for many Indo-European Romance languages, but the template on its website does not lend itself to automatic data extraction tools. We can try to access this websites using GFCE, but case many times, the ratings assigned are sentences instead of a single label.

\item Another common reason is that on a number of these websites, the claim statements are not well-specified. Take for example Faktograf\footnote{\url{https://faktograf.hr/ocjena-tocnosti/}}, a website performing fact-checking in Croatian. On this website
% \footnote{\url{https://faktograf.hr/2021/01/28/zakon-o-radu-ne-regulira-beneficirani-radni-staz-kao-sto-to-navodi-vili-beros/}}
, we can neither properly extract the claim statements nor do they clearly mention the rating assigned to the articles. 

\item For a small percentage of the claim statements, Google search did not yield any results. We omitted all of these claims from our training, development, and test sets. These are only a very small percentage of claims, so we remove them from all models.

\end{enumerate}

Because of these reasons, a large number of websites in a number of languages could not be crawled.

There are two ways we obtain our claims, labels, and other metadata. One is the Google's Fact Check Explorer (GFCE)\footnote{\url{https://toolbox.google.com/factcheck/explorer}}, and the other is by crawling from the respective fact-checking website. In case, the links are available on GFCE, we download other metadata by visiting the website. Also, we will release the label mapping we created along with the dataset.~\Cref{tab:languages} provides more details on the dataset we collected.

% Like mentioned in the main paper, we exclude a number of websites where a single label dominates in the dataset. 

% We provide more details on construction of out-of-domain (\alphaOne) test set. 
\begin{table}[ht]
\centering
\resizebox{\columnwidth}{!}{%
\begin{tabular}{lcc}
\toprule
Dataset             & Model           & RunTime \\ \midrule
\datasetName        & Claim           & 1.5 hr  \\
\datasetName        & Claim+Meta      & 1.5 hr  \\
\datasetName        & Attn-EA        & 2.3 hr  \\
\datasetName        & Attn-EA + Meta & 2.3 hr  \\
\datasetName + Eng & Claim+Meta      & 2.5 Hr  \\ 
  \datasetName + Eng & Attn-EA + Meta & 4.1 Hr  \\
  \bottomrule
\end{tabular}
}
\caption{Average Training time of the models trained}
\label{tab:runtime}
\end{table}

%%% Local Variables:
%%% mode: latex
%%% TeX-master: "../../acl2021"
%%% End:

% Please add the following required packages to your document preamble:
% \usepackage{booktabs}
% \usepackage[table,xcdraw]{xcolor}
% If you use beamer only pass "xcolor=table" option, i.e. \documentclass[xcolor=table]{beamer}
% \usepackage[normalem]{ulem}
% \useunder{\uline}{\ul}{}
\begin{table*}[ht]
% \resizebox{\textwidth}{!}{%
\begin{tabular}{@{}lllllllll@{}}
\toprule
Language    & \begin{tabular}[c]{@{}l@{}}ISO\\ 639-1\\ code\end{tabular} & FactChecker                                                   & Language Family & Train                 & Dev                   & \alphaOne & \alphaTwo & \alphaThree \\ \midrule
Arabic      & ar                                                         & misbar.com                                                    & Afro-Asiatic    & \cmark & \cmark & \cmark    &                          &                            \\
Bengali     & bn                                                         & dailyo.in                                                     & IE: Indo-Aryan  &                       &                       &                          &                          & \cmark      \\
Spanish     & es                                                         & chequeado.com                                                 & IE: Romance     & \cmark & \cmark & \cmark    &                          &                            \\
Persian     & fa                                                         & factnameh.com                                                 & IE: Iranian     &                       &                       &                          &                          & \cmark      \\
Indonesian  & id                                                         & cekfakta.com                                                  & Austronesian    & \cmark & \cmark & \cmark    &                          &                            \\
Indonesian  & id                                                         & cekfakta.tempo.co                                             & Austronesian    &                       &                       &                          & \cmark    &                            \\
Italian     & it                                                         & pagellapolitica.it                                            & IE: Romance     & \cmark & \cmark & \cmark    &                          &                            \\
Italian     & it                                                         & agi.it                                                        & IE: Romance     &                       &                       &                          & \cmark    &                            \\
Hindi       & hi                                                         & aajtak.in                           & IE: Indo-Aryan  & \cmark & \cmark & \cmark    &                          &                            \\
Hindi       & hi                                                         & hindi.newschecker.in               & IE: Indo-Aryan  &                       &                       &                          & \cmark    &                            \\
Gujarati    & gu                                                         & gujarati.newschecker.in             & IE: Indo-Aryan  &                       &                       &                          &                          & \cmark      \\
Georgian     & ka                                                         & factcheck.ge                                                        & Kartvelian     &\cmark & \cmark & \cmark    &                            \\
Marathi     & mr                                                         & marathi.newschecker.in              & IE: Indo-Aryan  &                       &                       &                          &                          & \cmark      \\
Punjabi     & pa                                                         & punjabi.newschecker.in.txt                                    & IE: Indo-Aryan  &                       &                       &                          &                          & \cmark      \\
Polish      & pl                                                         & demagog.org.pl & IE: Slavic      & \cmark & \cmark & \cmark    &                          &                            \\
Portuguese  & pt                                                         & piaui.folha.uol.com.br                                        & IE: Romance     & \cmark & \cmark & \cmark    &                          &                            \\
Portuguese  & pt                                                         & poligrafo.sapo.pt                                             & IE: Romance     & \cmark & \cmark & \cmark    &                          &                            \\
Romanian    & ro                                                         & factual.ro                                                    & IE: Romance     & \cmark & \cmark & \cmark    &                          &                            \\
Norwegian   & no                                                         & faktisk.no                          & IE: Germanic    &                       &                       &                          &                          & \cmark      \\
Sinhala     & si                                                         & srilanka.factcrescendo.com                                    & IE              &                       &                       &                          &                          & \cmark      \\
Serbian     & sr                                                         & istinomer.rs                                                  & IE: Slavic      & \cmark & \cmark & \cmark    &                          &                            \\
Tamil       & ta                                                         & youturn.in                          & Dravidian       & \cmark & \cmark & \cmark    &                          &                            \\
Albanian    & sq                                                         & kallxo.com                          & IE: Albanian    &                       &                       &                          &                          & \cmark      \\
Albanian    & sq                                                         & faktoje.al      & IE: Albanian    &                       &                       &                          &                          & \cmark      \\
Russian     & ru                                                         & factcheck.kz                        & IE: Slavic      &                       &                       &                          &                          & \cmark      \\
Turkish     & tr                                                         & dogrulukpayi.com                    & Turkic          & \cmark & \cmark & \cmark    &                          &                            \\
Turkish     & tr                                                         & teyit.org                           & Turkic          &                       &                       &                          & \cmark    &                            \\
Azerbaijani & az                                                         & faktyoxla.info                      & Turkic          &                       &                       &                          &                          & \cmark      \\
Portuguese  & pt                                                         & aosfatos.org                        & IE: Romance     &                       &                       &                          &                          & \cmark      \\
German      & de                                                         & correctiv.org                       & IE: Germanic    & \cmark & \cmark & \cmark    &                          &                            \\
Dutch       & nl                                                         & nieuwscheckers.nl                   & IE: Germanic    &                       &                       &                          &                          & \cmark      \\
French      & fr                                                         & fr.africacheck.org                  & IE: Romance     &                       &                       &                          &                          & \cmark      \\ \bottomrule
\end{tabular}
% }
\label{tab:languages}
\caption{Details of the \datasetName\, dataset. Our dataset belongs to 25 typologically diverse languages across 11 language families. The table shows the composition of training, development, and three challenge sets. IE: denotes Indo-Aryan}
\end{table*}

\section{Reproducibility}
In this section, we provide details on our hyperparameter settings along with some comments on reproducibility.

\subsection{Models and Code} As described in the main paper, we used multilingual BERT for performing our experiments. 
% We used the base version of mBERT from the huggingface repository. 
We implemented all our models in PyTorch using the transformers library~\citep{wolf2019huggingface}. 

% \paragraph{Hyperparameters} 

\subsection{Computing Infrastructure Used}
All of our experiments required access to GPU accelerators. We ran our experiments on three machines: Nvidia Tesla V100 (16 GB VRAM), Nvidia Tesla P100 (16 GB VRAM), Tesla A100 (40 GB VRAM). Our experiments for the claim-only model were run on V100, and P100 GPUs and evidence-based models required larger VRAM, so they were run on A100 GPUs. 

\subsection{Hyperparameters and Fine-tuning Details}
\begin{enumerate}

\item  We used the mBERT-\textit{base} model for all of our experiments. This model has 12 layers each with hiddem size of 768 and number of attention heads equal to 12. Total number of parameters in this model is 125 million. We set all the hyper-parameters as suggested by~\citet{devlin-etal-2019-bert}, except the batch size which is fixed to 8.

\item  All our models were run with four random seeds (seed $= [1,2,3,4]$) and the numbers reported in paper are the means of these four runs. 
We fine-tuned all models for ten epochs and the model performing the best on development set across all epochs was chosen as the final model.

\item  Due to constraints on the VRAM of the GPUs, we restricted the number of evidence documents to five. 
    
\end{enumerate}

\paragraph{Average Run times} Average training times are presented in \cref{tab:runtime}.

\end{document}